\documentclass[letterpaper, 10 pt, conference]{ieeeconf}  

\IEEEoverridecommandlockouts                              

\usepackage[linesnumbered,ruled,vlined]{algorithm2e}

\SetKwInput{KwInput}{Params}                
\SetKwInput{KwOutput}{Output}
\SetKw{KwReturn}{Return}
\SetKwFunction{KwFunction}{pGSCAM}

\usepackage{hyperref}
\usepackage[pdftex]{graphicx}
\usepackage{epstopdf}
\usepackage{authblk}
\AppendGraphicsExtensions{.tif}
\usepackage{cite}
\usepackage{amsmath}

\title{\LARGE \bf
Towards Explainable LiDAR Point Cloud Semantic Segmentation via Gradient Based Target Localization
}

\author{Abhishek Kuriyal$^{1}$ and Vaibhav Kumar$^{1}$

\thanks{$^{1}$Abhishek Kuriyal and Vaibhav Kumar are with the \href{https://sites.google.com/iiserb.ac.in/geoai/home?authuser=0}{GeoAI4Cities Lab}, Department of Data Science and Engineering, IISER Bhopal.\\
\tt\small \{{abhishek19, vaibhav\}}@iiserb.ac.in}

}

\begin{document}

\maketitle
\thispagestyle{empty}
\pagestyle{empty}

\begin{abstract}

Semantic Segmentation (SS) of LiDAR point clouds is essential for many applications, such as urban planning and autonomous driving. While much progress has been made in interpreting SS predictions for images, interpreting point cloud SS predictions remains a challenge. This paper introduces pGS-CAM, a novel gradient-based method for generating saliency maps in neural network activation layers. Inspired by Grad-CAM, which uses gradients to highlight local importance, pGS-CAM is robust and effective on a variety of datasets (SemanticKITTI, Paris-Lille3D, DALES) and 3D deep learning architectures (KPConv, RandLANet). Our experiments show that pGS-CAM effectively accentuates the feature learning in intermediate activations of SS architectures by highlighting the contribution of each point. This allows us to better understand how SS models make their predictions and identify potential areas for improvement. Relevant codes are available at \href{https://github.com/geoai4cities/pGS-CAM}{https://github.com/geoai4cities/pGS-CAM}.

\end{abstract}

\section{INTRODUCTION}

 Deep Learning (DL) techniques, such as Convolutional Neural Networks (CNNs) employed for feature extraction, have gained substantial traction in various applications, including instance classification, semantic segmentation, object detection, and visual tracking \cite{chai2021deep, dong2021survey}. Nonetheless, a significant challenge lies in the interpretability of their outcomes.
This challenge can benefit from the effective use of various Explainable AI (XAI) methods, with extensive resources available in recent reviews by Van et al. \cite{van2022explainable} and Linardatos et al. \cite{linardatos2020explainable}.
 Over the years, Grad-CAM  \cite{selvaraju2017grad} has been one of the most widely used techniques that uses the gradients of the output with respect to the final convolutional layer to produce localization maps (saliency maps) highlighting important regions. These important regions are most influential in the model's decision-making process. However, the implementation of Grad-CAM is predominantly observed in studies focusing on tasks related to images.
\begin{figure}[h]
\centering
\includegraphics[width=3.0in]{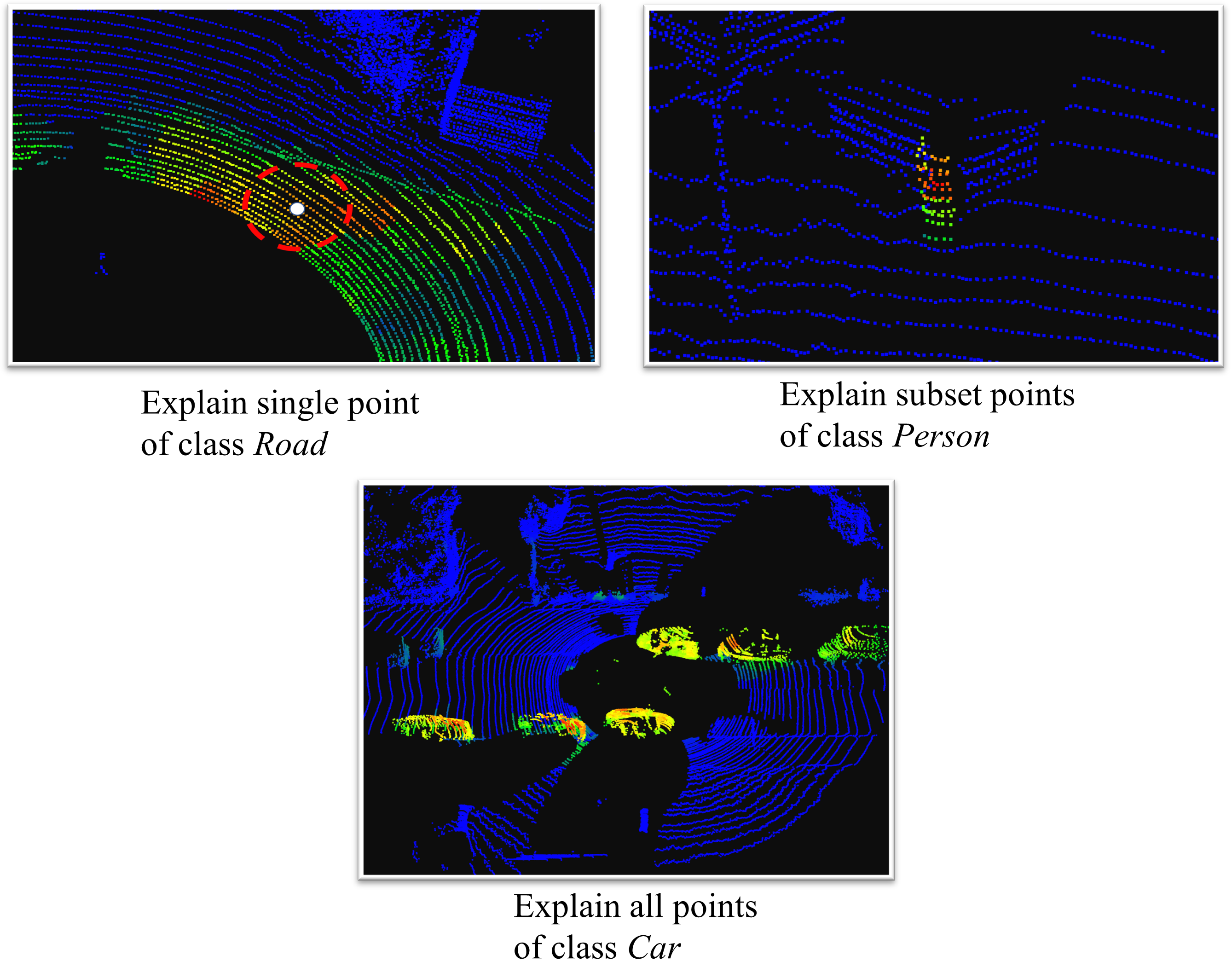}
\caption{pGS-CAM for single point (white dot) and class \textit{Road}, points corresponding to subset of class \textit{Person}, points corresponding to all points of class \textit{Car}. All heatmaps are obtained with respect to the final activation layers.}
\label{subset_pgscam}
\end{figure}

In the past few years, there has been a notable rise in interest towards Semantic Segmentation (SS) applied to Light Detection and Ranging (LiDAR) point clouds. This trend is attributed to LiDAR's capacity for precise 3D environmental perception. The reader can refer to the following review papers \cite{dong2021survey, guo2020deep, vinodkumar2023survey} highlighting deep learning-based point cloud semantic segmentation research in recent years.



Recently, efforts are made to explain feature extraction of point cloud DL architectures for instance classification and object detection. 
Zhang et al. \cite{zhang2019explaining} explains PointNet by visualizing what each point function has learnt followed by generating class-attentive response maps. However, their approach involved modifying the network's architecture to utilize class-attentive features. Zheng et al. \cite{zheng2019pointcloud} proposed a differentiable point-shifting process that computes contribution scores to input points based on the loss value, but it sacrifices global interpretability for highly accurate individual point contributions. The resulting heatmaps resemble fine details similar to LRP \cite{binder2016layer} methods used in images. 
Other methods such as \cite{souai2022deep, dworak2022adaptation} utilize Grad-CAM for instance classification and object detection of point clouds, respectively.
Needless to say, there is a lack of dedicated research on interpretable SS of point clouds. Additionally, no existing study has proposed and implemented a method utilizing gradients in architectures for interpretation of large-scale outdoor point segmentation. Our study aims to fill these gaps by making the following major contributions:
\begin{itemize}
\item We propose pGS-CAM (point Grad-Seg Class Activation Mapping), a novel method inspired by Grad-CAM. It creates saliency maps for each activation layer in LiDAR Semantic Segmentation (LSS) architectures, revealing influential points. Our approach marks the first to offer visual explanations for LSS process in 3D DL architectures.

\item  Heatmaps generated via pGS-CAM can provide valuable purpose in comprehending architecture's working. One such utility is to verify that the network does not fixate on peculiar intricacies unique to the training scans, which may not extend to unseen scans. 

\item We provide comprehensive qualitative analysis on two well-known point-based architectures, KPConv \cite{thomas2019kpconv} and RandLA-Net \cite{hu2020randla}. We utilize widely recognized MLS datasets, including SemanticKITTI \cite{behley2019SemanticKITTI}, Paris-Lille3D \cite{thomas2019kpconv}, and the DALES ALS dataset \cite{varney2020DALES}. We emphasize distinctions in the inner workings of architectures, present counterfactual explanations, provide quantitative analysis of pGS-CAM explanations etc., to prove method's robustness.
\end{itemize}

\section{Methodology}

Let \({\{A^{k}\}}^K_{k=1}\) denote the chosen feature maps of interest, representing the K kernels from the final convolutional layer of the classification network, and \(l^{c}\) represents the logit for a selected class \(c\). Grad-CAM calculates the importance weight \(\alpha^{k}_{c}\) for each feature map \(A^{k}\) by averaging the gradients of \(l^{c}\) with respect to all \(N\) pixels (indexed as \(u, v\)) (see Equation \ref{gradcam_weights}).
The heatmap 
\begin{equation}
\label{gradcam_heatmap}
L^c = ReLU\left(\sum_{k} \alpha^{k}_{c} A^{k}\right)
\end{equation}
with
\begin{equation}
\label{gradcam_weights}
\alpha^{k}_{c} = \frac{1}{N}\sum_{u,v} \frac{\partial l^{c}}{\partial A^{k}_{uv}}
\end{equation}

is formed by aggregating the feature maps using weights \(\alpha^{k}_{c}\). This summation is followed by the application of the ReLU function at the pixel level to zero out negative values, thus emphasizing areas that positively contribute to the classification decision for class \(c\).

A fundamental challenge arises from the distinction that, while a image classification network generates a single logit \(l^{c}\) for each input image, a LSS network provides logits \(l^{c}_{i}\) for each of the \(N\) points within the point cloud \(P\). For any intermediate activation layer \(A\) in \(R^{(M\times k)}\) (consisting of M downsampled points with k feature dimensions), the gradient of the logit \(l_i\) for the target class \(c\) with respect to \(A_k\) is expressed by Equation \ref{partialLogit}.


\begin{equation}
\label{partialLogit}
\frac{\partial l^{c}_{i}}{\partial A^{k}}
\end{equation}

To capture the collective gradient impact across all logits \(l^{c}_{i}\), where \(i \in N\), we employ an aggregation operation. Our observations indicate that summation is particularly effective for this purpose. Furthermore, there may be cases where we require explanations for a limited subset of points, rather than the entire point cloud. To address this need, we introduce pGS-CAM, which focuses on a subset of points denoted as \(P'\), consisting of \(N'\) points. We aim to provide individual explanations for each point within this subset. \(P'\) may encompass a single point, points associated with a specific class instance, or all \(N\) points within the point cloud (refer to Fig. \ref{subset_pgscam} for a visual demonstration). As a result, the overall pointwise gradient influence for this particular subset is defined by Equation \ref{overallInfluence}.

\begin{equation}
\label{overallInfluence}
\frac{\partial l^{c}_{1}}{\partial A^{k}} + \frac{\partial l^{c}_{2}}{\partial A^{k}} + ... + \frac{\partial l^{c}_{N'}}{\partial A^{k}} = \frac{\partial \sum_{i=1}^{N'}l_{i}^{c}}{\partial A^{k}} = \frac{\partial L_{P'}}{\partial A^{k}}
\end{equation}

Thereafter, the gradient influence \(G^k\) similar to \(\alpha^{k}_{c}\) in Grad-CAM is approximated. \(G^k\) (see Equation \ref{Gk}) is calculated by aggregating over M points in activation layer \(A_k\).

\begin{equation}
\label{Gk}
G^{k} = \sum_{j=1}^{M}\left[ 
\frac{\partial L_{P'}}{\partial A^{k}_{j}}
\right]
\end{equation}

Heatmap at activation layer \(A^{k}\) is generated via matrix multiplication of \(G^k\) with \(A^k\) followed by ReLU operation to highlight positive contributions only and MinMax normalization to align floating point values between 0 and 1 (see Equation \ref{heatmap}). The normalization is done to maintain heatmap consistency over multiple point cloud instances aiding qualitative and quantitative comparisons. 

\begin{equation}
\label{heatmap}
H = MinMax(ReLU(\sum_{k} G^{k}A^{k}))
\end{equation}

The standard pGS-CAM produces saliency maps for each neural network activation layer. However, in many semantic segmentation architectures with Encoder-Decoder structures, the resolution of these maps decreases as we move through the layers. To overcome this, we propose KDTree upsampling to maintain the original point cloud resolution (see Algorithm \ref{pgscam}). \textbf{Note} that unlike image classification networks, where intermediate activation maps can be used as coarse heatmaps, this isn't the case for LSS networks. Localization is apparent only after applying gradient influence \(G_{k}\), highlighting a significant difference between these architectural approaches.


  


\begin{algorithm}[!ht]
\caption{pGS-CAM algorithm}\label{pgscam}
\DontPrintSemicolon
\SetKwFunction{FMain}{Main}
\SetKwProg{Fn}{Function}{:}{}
\Fn{\FMain{Point cloud $P$, Segmentation Network $f$}}{
  \KwInput {Point cloud $P$, Segmentation Network $f$}
  \KwData{Dataset $D$ with $N$ points}
  \tcp{\(P^{M}\): Point resolution with resolution \(M\)}
  \tcp{\(P^{N}\): Point resolution with resolution \(N\)}
  $l^{c} \gets f(P)$\;
  $G^{k} \gets \sum_{j=1}^{M}\left[\frac{\partial L_{P'}}{\partial A^{k}_{j}}\right]$\;
  $H_{A} \gets \text{MinMax}(\text{ReLU}(\sum_{k} G^{k}A^{k}))$\;
  tree $\gets$ KDTree($P^{M}$, leaf\_size=40)\;
  upscale\_idx $\gets$ tree.query($P^{N}$)\;
  $H_{A} \gets H_{A}$[upscale\_idx]\;
  \KwRet{$H_{A}$}\;
}
\end{algorithm}

\section{Experimentation and Discussion}
 The subsequent sections will illustrate the robustness of pGS-CAM. In Section \textbf{A}, we offer a qualitative analysis of saliency maps and compare how these architectures function at their core. In Section \textbf{B}, we connect pGS-CAM heatmaps with dimensionality reduction techniques like PCA and t-SNE to validate the heatmap results and offer diverse perspectives for heatmap analysis. Section \textbf{C} presents intriguing counterfactual explanations, highlighting points that significantly influence the network's performance. Lastly, in Section \textbf{D}, we provide a quantitative analysis of pGS-CAM explanations through high-drop and low-drop experiments.
\subsection{Qualitative Analysis of Saliency Maps}
\begin{figure*}[h]
\centering
\includegraphics[width=5.0in]{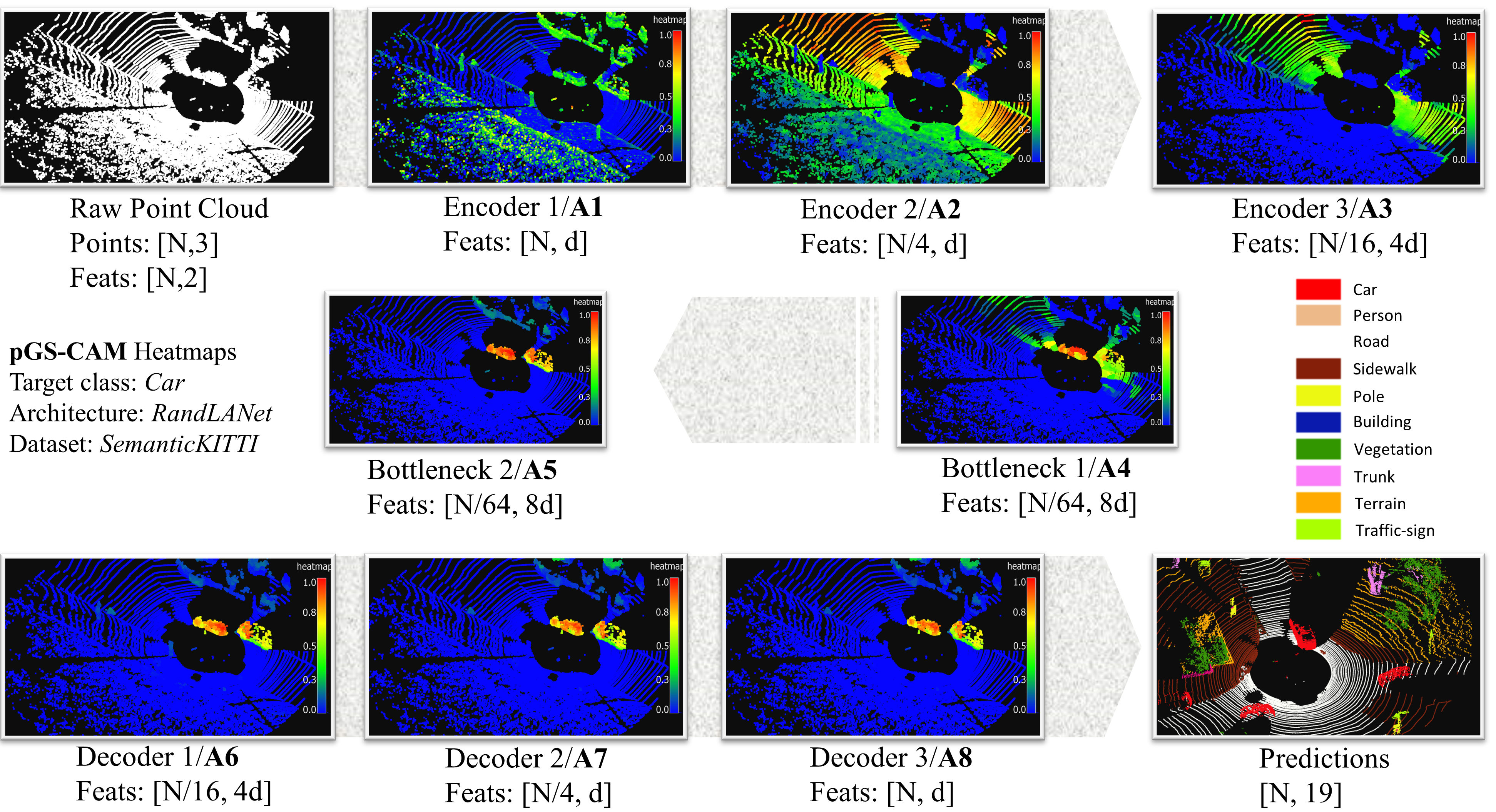}
\caption{pGS-CAM heatmaps for SemanticKITTI \textit{car} class and RandLA-Net architecture.}
\label{full_kitti}
\end{figure*}

Fig. \ref{full_kitti} demonstrates pGS-CAM explanations for the car class in the SemanticKITTI dataset using the RandLA-Net architecture. In the absence of KDTree upsampling, the heatmaps exhibit varying resolutions due to differences in activation layer resolutions. This is particularly challenging in bottleneck layers with reduced point density. To address this issue, Fig. \ref{full_kitti} presents full-resolution heatmaps using Algorithm \ref{pgscam} and nearest neighbor upsampling, offering a close visual representation of decision-making at each layer. A consistent pattern emerges, with edge-like structures in the initial activation layer A1 and localization starting from A5 onwards. A2-A4 heatmaps highlight road and terrain points, which influence car class predictions, and A8 closely resembles per-point predictions for the car class. Localization generally begins around A5 in RandLA-Net.

\begin{figure*}[h]
\centering
\includegraphics[width=5.0in]{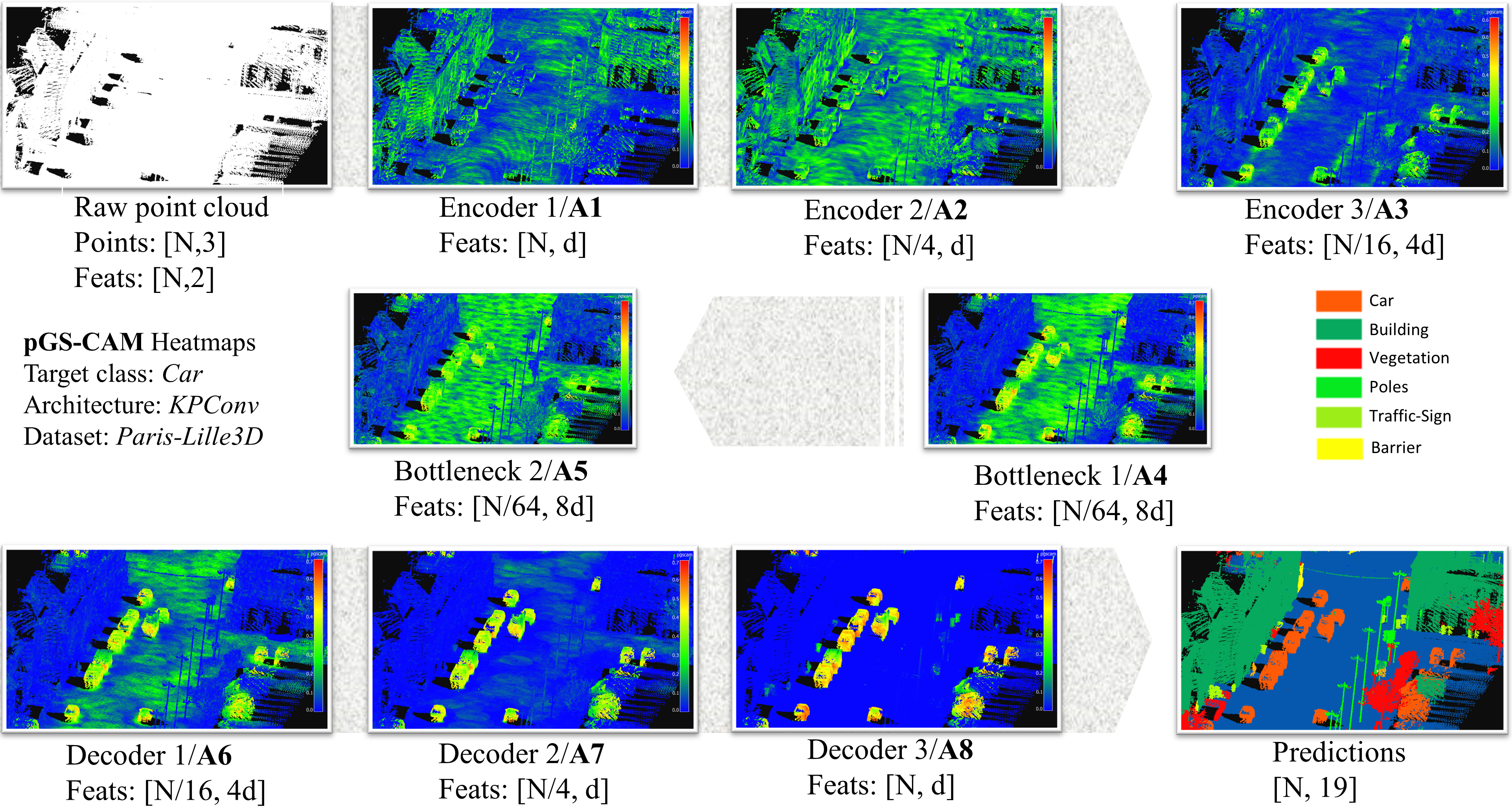}
\caption{pGS-CAM heatmaps for Paris-Lille3D \textit{car} class and KPConv architecture.}
\label{paris_lille3d}
\end{figure*}

\begin{figure*}[h]
\centering
\includegraphics[width=5.0in]{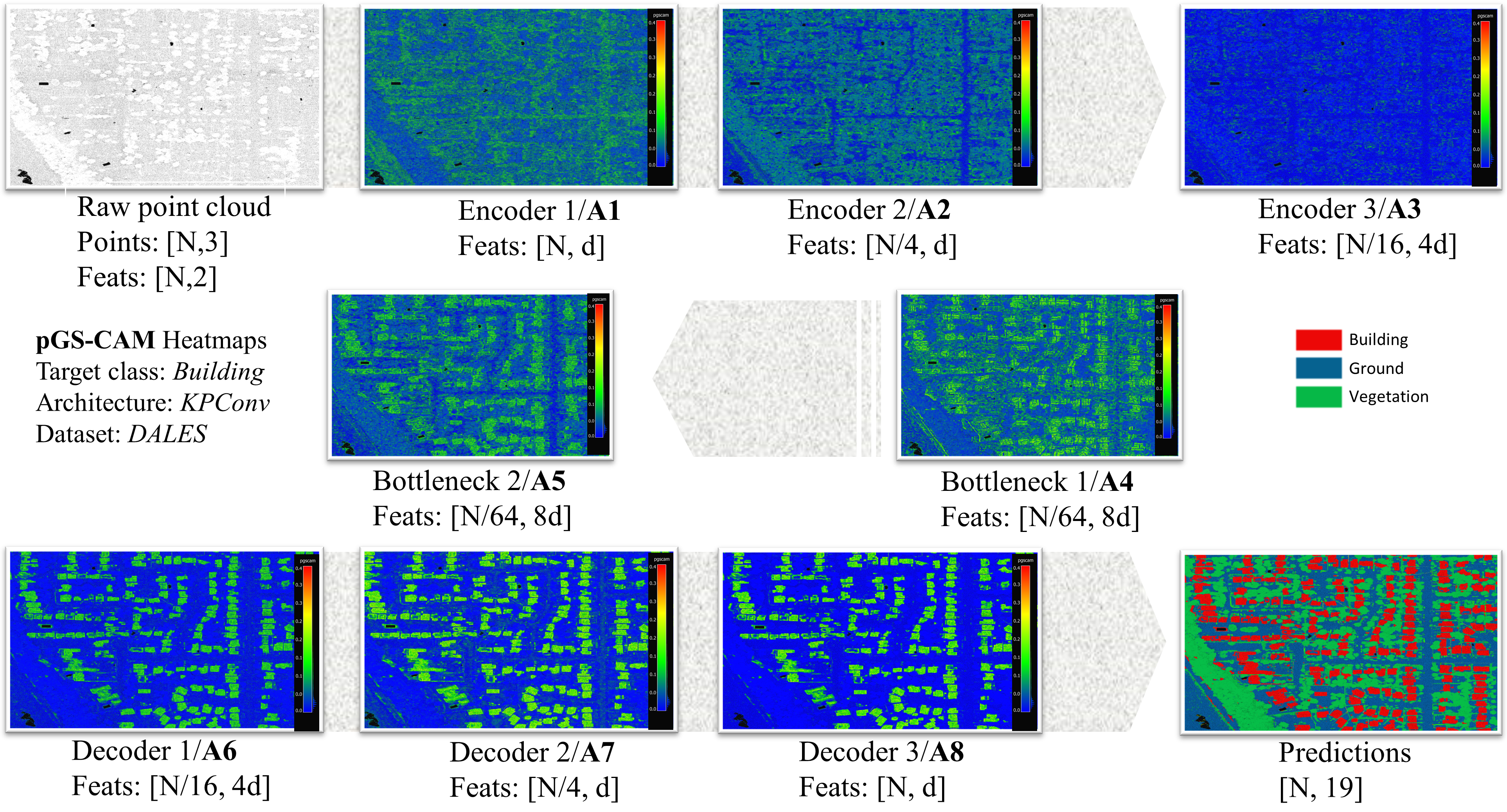}
\caption{pGS-CAM heatmaps for DALES \textit{building} class and KPConv architecture.}
\label{DALES_heatmaps}
\end{figure*}

Fig. \ref{paris_lille3d} displays pGS-CAM explanations for the KPConv architecture for the Paris-Lille3D dataset. Similar to Fig. \ref{full_kitti}, initial activation layers A1-A2 show edge-like structures. However, localization in KPConv starts from the A3 layer, indicating its superior representation capabilities. Fig. \ref{DALES_heatmaps} presents pGS-CAM heatmaps for the KPConv architecture for the DALES aerial dataset. The pattern is similar to the Paris-Lille3D dataset, with early localization in the encoder layers and potential redundancy in later layers (A6, A7, and A8). Poor gradient highlights at encoder 3 suggest a high number of negative inputs affecting information representation in that layer.

It is important to highlight that pGS-CAM heatmaps consistently exhibit reproducibility across all datasets and architectures, making them valuable for benchmarking purposes.


\subsection{Interpreting pGS-CAM via Dimensionality Reduction}
\begin{figure}[!t]
\centering
\includegraphics[width=3.3in]{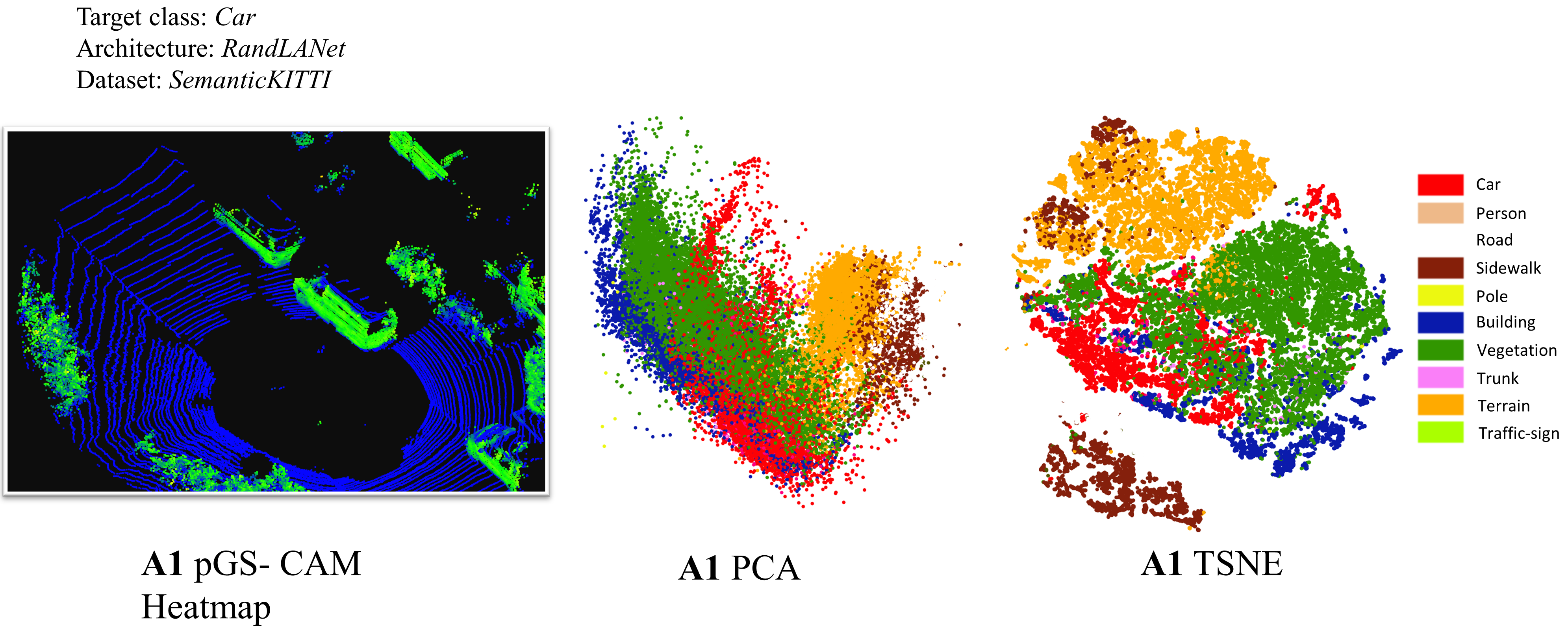}
\caption{pGS-CAM heatmaps comparison with PCA and TSNE plots for initial activation layer and target class \(car\). Notice the highlights of vegetation points in heatmap which is reflected by the cluttering of car and vegetation in PCA and TSNE plots.}
\label{pgscam_act0}
\end{figure}

\begin{figure}[!t]
\centering
\includegraphics[width=3.3in]{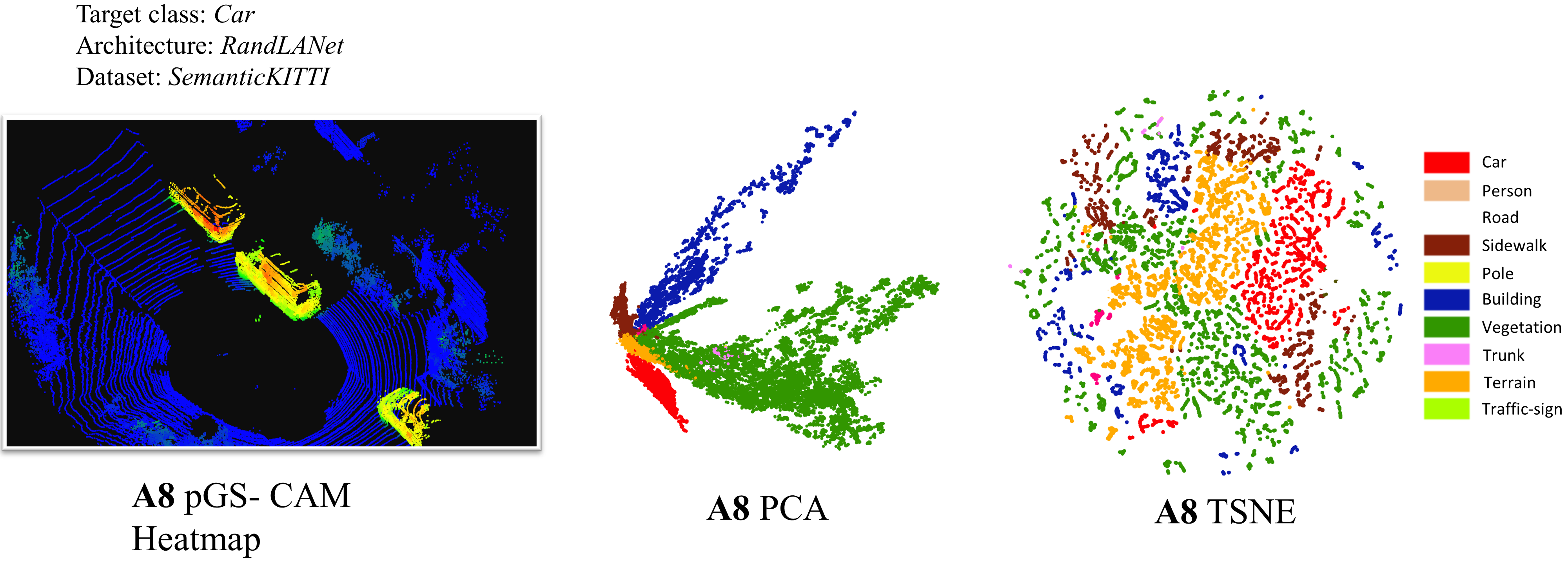}
\caption{pGS-CAM heatmaps comparison with PCA and TSNE plots for final activation layer. Notice the presence of highlighted points corresponding to the "vegetation" class in close proximity to points belonging to the "car" class within the heatmap. This finding is further supported by the t-SNE plot, which demonstrates a slight overlap between these two classes.}
\label{pgscam_actf}
\end{figure}

We conducted a comparative analysis between pGS-CAM heatmaps and 2D dimensionality reduction techniques applied to corresponding activation layers. Specifically, we employed Principal Component Analysis (PCA) and t-Distributed Stochastic Neighbor Embedding (t-SNE). In the initial activation layer, pGS-CAM revealed the presence of clustered points and overlapping features among car, vegetation, and building points across all three methods. However, in terms of interpretability and result differentiation, t-SNE exhibited more intuitive and distinct patterns when compared to PCA. This superiority is attributed to t-SNE's advanced dimensionality reduction capabilities (see Fig. \ref{pgscam_act0}). Transitioning to the final activation layer, we observed a localized pGS-CAM heatmap, consistent with the PCA and t-SNE plots (see Figure \ref{pgscam_actf}). In a specific case, we observed a slight highlighting of vegetation points alongside car points in the heatmap, aligning with the t-SNE plots, where red points (representing cars) were surrounded by green points (representing vegetation). These findings highlight the coherence between pGS-CAM and PCA, t-SNE observations and suggest that pGS-CAM offers more intuitive per-point explanations compared to both techniques.

\subsection{Counterfactual pGS-CAM Explanations}
An intriguing analysis, applicable to any explanation method, involves identifying points that have a negative impact on the performance of a specific class. This valuable insight can be obtained through a simple adjustment in the gradient influence \(G_{k}\). By negating the gradient influence (as described in Equation \ref{counterfactual}), we can spotlight regions that adversely affect the decision of the target class.

\begin{equation}
\label{counterfactual}
G^{k} = \sum_{j=1}^{M}\left[- 
\frac{\partial L_{P'}}{\partial A^{k}_{j}}
\right]
\end{equation}

\begin{figure}[!t]
\centering
\includegraphics[width=3.3in]{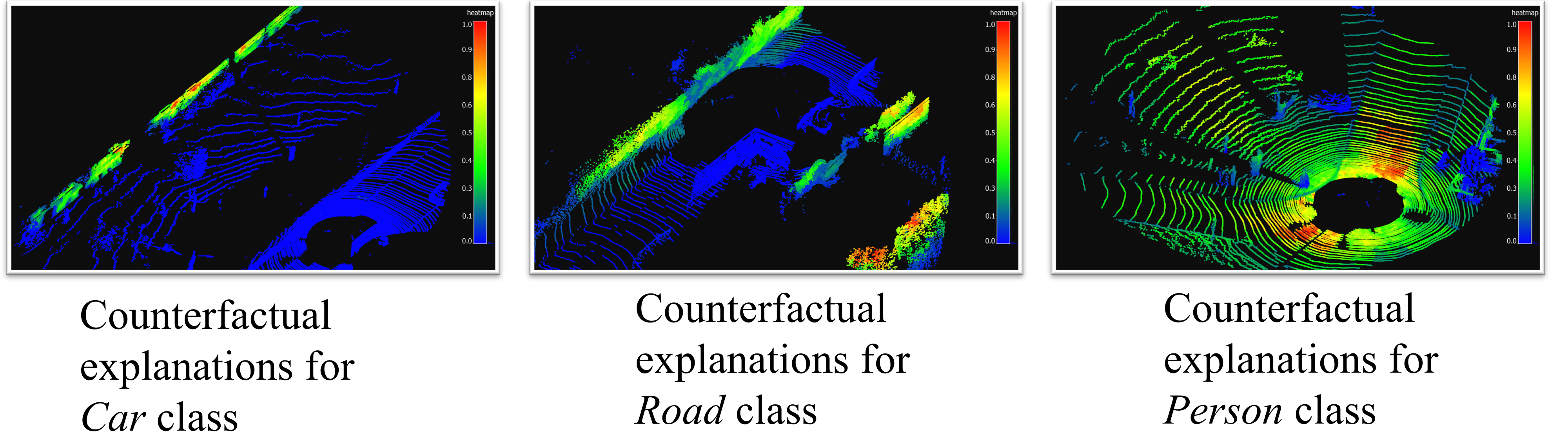}
\caption{Using negative gradient influence to highlight counterfactual pGS-CAM explanations. These findings offer insightful inferences into the contributing factors behind the adverse performance of the target class.}
\label{counterfactual_pgscam}
\end{figure}

Fig. \ref{counterfactual_pgscam} provides illustrative examples of these scenarios. The first example presents a counterfactual explanation for the car class, suggesting that the removal of fence points might enhance the performance of this class. Similarly, the second example yields a similar result for the road class, highlighting the influence of vegetation points. However, the most intriguing case is the third example, which suggests that the performance of the person class could be improved by eliminating all ground points. This observation aligns with intuition, as LSS architectures often face challenges when dealing with substantial differences in object sizes. Approaches such as KPConv employ adaptive kernel convolutions to address these peculiarities.

\subsection{Quantitative Point-Drop Experiments}

\begin{figure}[!t]
\centering
\includegraphics[width=3.3in]{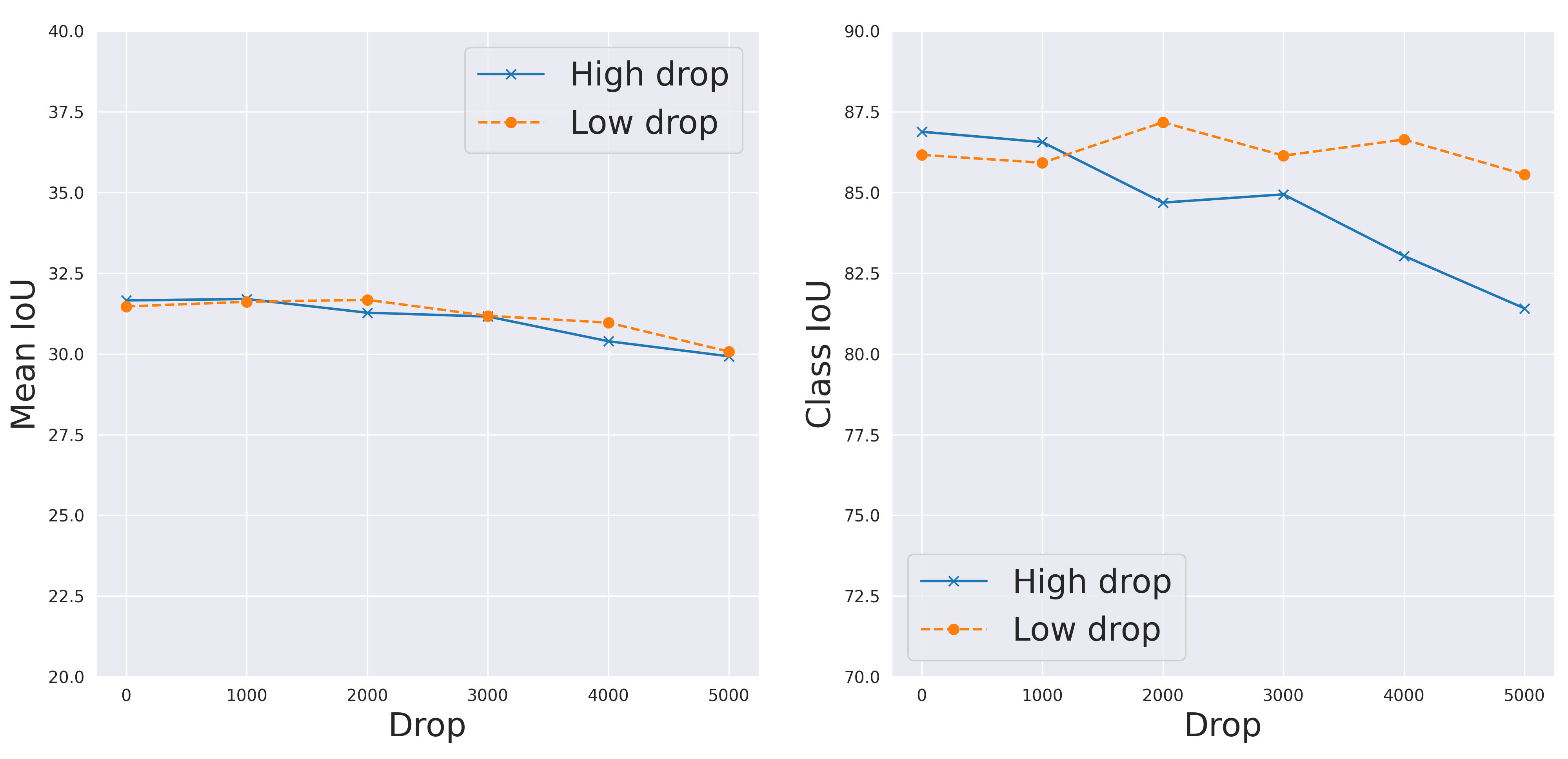}
\caption{Quantitative Point-Drop removal for \textit{car} class and affect on mean IoU performance for all class. Notice a significant decline in the performance of the \textit{car} class following a high-drop scenario.}
\label{point_drop_plot}
\end{figure}

Point Drop serves as a robustness assessment method for explanation techniques. It involves removing points with high saliency scores (high drop), causing a significant performance drop, or dropping points with low saliency scores (low drop), resulting in a minor performance decrease. In our experiments, we computed class-specific Intersection over Union (IoU) stability and mean IoU stability after each point drop attack. Interestingly, in certain scenarios, the model correctly identified objects as "car" even after the removal of 5000 points. Fig. \ref{point_drop_plot} illustrates the IoU difference after each drop attack, averaged across point clouds from the SemanticKITTI validation set. High drop attacks led to a substantial class-specific IoU decrease, averaging at 1.36, while low drop attacks maintained relatively stable class-specific IoU, with an average decrease of 0.15. This underscores the effectiveness of pGS-CAM. When calculating mean IoU for all classes, not limited to the target class, the model's performance remained reasonably stable. The average decrease in mean IoU was 0.43 for high drop attacks and 0.34 for low drop attacks. Consequently, the overall performance decline across all classes was less severe compared to focusing solely on the target class.


\section{Conclusion}

We introduced pGS-CAM, a gradient based approach for explaining point cloud semantic segmentation models. This method reveals discernible patterns across network layers and consistently performs on popular datasets. When comparing architectures, we found differences in localization initiation, with KPConv displaying early localization compared to RandLA-Net. pGS-CAM effectively aligns with 2D dimensionality reduction techniques, highlighting class overlaps. We also proposed a gradient influence modification for counterfactual explanations that improved class-specific performance. Our Point-Drop experiment established pGS-CAM's robustness, making it a valuable benchmark. In summary, pGS-CAM provides fresh insights into the decision making of point cloud semantic segmentation models.







\bibliographystyle{IEEEtran}
\bibliography{main}

\begin{thebibliography}{10}
\providecommand{\url}[1]{#1}
\csname url@samestyle\endcsname
\providecommand{\newblock}{\relax}
\providecommand{\bibinfo}[2]{#2}
\providecommand{\BIBentrySTDinterwordspacing}{\spaceskip=0pt\relax}
\providecommand{\BIBentryALTinterwordstretchfactor}{4}
\providecommand{\BIBentryALTinterwordspacing}{\spaceskip=\fontdimen2\font plus
\BIBentryALTinterwordstretchfactor\fontdimen3\font minus
  \fontdimen4\font\relax}
\providecommand{\BIBforeignlanguage}[2]{{%
\expandafter\ifx\csname l@#1\endcsname\relax
\typeout{** WARNING: IEEEtran.bst: No hyphenation pattern has been}%
\typeout{** loaded for the language `#1'. Using the pattern for}%
\typeout{** the default language instead.}%
\else
\language=\csname l@#1\endcsname
\fi
#2}}
\providecommand{\BIBdecl}{\relax}
\BIBdecl

\bibitem{chai2021deep}
J.~Chai, H.~Zeng, A.~Li, and E.~W. Ngai, ``Deep learning in computer vision: A
  critical review of emerging techniques and application scenarios,''
  \emph{Machine Learning with Applications}, vol.~6, p. 100134, 2021.

\bibitem{dong2021survey}
S.~Dong, P.~Wang, and K.~Abbas, ``A survey on deep learning and its
  applications,'' \emph{Computer Science Review}, vol.~40, p. 100379, 2021.

\bibitem{van2022explainable}
B.~H. Van~der Velden, H.~J. Kuijf, K.~G. Gilhuijs, and M.~A. Viergever,
  ``Explainable artificial intelligence (xai) in deep learning-based medical
  image analysis,'' \emph{Medical Image Analysis}, vol.~79, p. 102470, 2022.

\bibitem{linardatos2020explainable}
P.~Linardatos, V.~Papastefanopoulos, and S.~Kotsiantis, ``Explainable ai: A
  review of machine learning interpretability methods,'' \emph{Entropy},
  vol.~23, no.~1, p.~18, 2020.

\bibitem{selvaraju2017grad}
R.~R. Selvaraju, M.~Cogswell, A.~Das, R.~Vedantam, D.~Parikh, and D.~Batra,
  ``Grad-cam: Visual explanations from deep networks via gradient-based
  localization,'' in \emph{Proceedings of the IEEE international conference on
  computer vision}, 2017, pp. 618--626.

\bibitem{guo2020deep}
Y.~Guo, H.~Wang, Q.~Hu, H.~Liu, L.~Liu, and M.~Bennamoun, ``Deep learning for
  3d point clouds: A survey,'' \emph{IEEE transactions on pattern analysis and
  machine intelligence}, vol.~43, no.~12, pp. 4338--4364, 2020.

\bibitem{vinodkumar2023survey}
P.~K. Vinodkumar, D.~Karabulut, E.~Avots, C.~Ozcinar, and G.~Anbarjafari, ``A
  survey on deep learning based segmentation, detection and classification for
  3d point clouds,'' \emph{Entropy}, vol.~25, no.~4, p. 635, 2023.

\bibitem{zhang2019explaining}
B.~Zhang, S.~Huang, W.~Shen, and Z.~Wei, ``Explaining the pointnet: What has
  been learned inside the pointnet?'' in \emph{CVPR Workshops}, 2019, pp.
  71--74.

\bibitem{zheng2019pointcloud}
T.~Zheng, C.~Chen, J.~Yuan, B.~Li, and K.~Ren, ``Pointcloud saliency maps,'' in
  \emph{Proceedings of the IEEE/CVF International Conference on Computer
  Vision}, 2019, pp. 1598--1606.

\bibitem{binder2016layer}
A.~Binder, S.~Bach, G.~Montavon, K.-R. M{\"u}ller, and W.~Samek, ``Layer-wise
  relevance propagation for deep neural network architectures,'' in
  \emph{Information science and applications (ICISA) 2016}.\hskip 1em plus
  0.5em minus 0.4em\relax Springer, 2016, pp. 913--922.

\bibitem{souai2022deep}
Y.~Souai, G.~Rouhafzay, and A.-M. Cretu, ``A deep-learning-based approach for
  saliency determination on point clouds,'' \emph{Engineering Proceedings},
  vol.~27, no.~1, p.~17, 2022.

\bibitem{dworak2022adaptation}
D.~Dworak and J.~Baranowski, ``Adaptation of grad-cam method to neural network
  architecture for lidar pointcloud object detection,'' \emph{Energies},
  vol.~15, no.~13, p. 4681, 2022.

\bibitem{thomas2019kpconv}
H.~Thomas, C.~R. Qi, J.-E. Deschaud, B.~Marcotegui, F.~Goulette, and L.~J.
  Guibas, ``Kpconv: Flexible and deformable convolution for point clouds,'' in
  \emph{Proceedings of the IEEE/CVF international conference on computer
  vision}, 2019, pp. 6411--6420.

\bibitem{hu2020randla}
Q.~Hu, B.~Yang, L.~Xie, S.~Rosa, Y.~Guo, Z.~Wang, N.~Trigoni, and A.~Markham,
  ``Randla-net: Efficient semantic segmentation of large-scale point clouds,''
  in \emph{Proceedings of the IEEE/CVF conference on computer vision and
  pattern recognition}, 2020, pp. 11\,108--11\,117.

\bibitem{behley2019SemanticKITTI}
J.~Behley, M.~Garbade, A.~Milioto, J.~Quenzel, S.~Behnke, C.~Stachniss, and
  J.~Gall, ``Semantickitti: A dataset for semantic scene understanding of lidar
  sequences,'' in \emph{Proceedings of the IEEE/CVF international conference on
  computer vision}, 2019, pp. 9297--9307.

\bibitem{varney2020DALES}
N.~Varney, V.~K. Asari, and Q.~Graehling, ``Dales: A large-scale aerial lidar
  data set for semantic segmentation,'' in \emph{Proceedings of the IEEE/CVF
  conference on computer vision and pattern recognition workshops}, 2020, pp.
  186--187.

\end{thebibliography}

\end{document}